\documentclass{article}

\usepackage{arxiv}

\usepackage[utf8]{inputenc} % allow utf-8 input
\usepackage[T1]{fontenc}    % use 8-bit T1 fonts
\usepackage{hyperref}       % hyperlinks
\usepackage{url}            % simple URL typesetting
\usepackage{booktabs}       % professional-quality tables
\usepackage{amsfonts}       % blackboard math symbols
\usepackage{nicefrac}       % compact symbols for 1/2, etc.
\usepackage{microtype}      % microtypography
\usepackage{lipsum}
\usepackage{graphicx}
\graphicspath{ {./images/} }

\usepackage{tikz}
\usepackage{amsmath}
\usepackage{algorithm}
\usepackage{algorithmic}
\usepackage{amsmath}
\usepackage{booktabs}
\usepackage{amsfonts}
\usepackage{graphicx}
\usepackage{float}
\usepackage{subcaption}
\usepackage{pgfplots}

\pgfplotsset{compat=1.18}
\usepackage{multirow}
\usepackage{tikz}
\usepackage{physics}
\usetikzlibrary{arrows.meta, positioning, fit, shapes.geometric, calc}
\usepackage{comment}
\usepackage{fontawesome5}
\usepackage{tabularx}
\newcommand{\targetmodel}{$\mathcal{T}$ }
\newcommand{\referencemodel}{$\mathcal{R}$ }

\newcommand{\publicdataset}{$\mathcal{O}$ }
\newcommand{\privatedataset}{$\mathcal{P}$ }
\newcommand{\threshold}{$\theta $ }
\usepackage{filecontents}

\author{
Faiz TALEB\\
    EDF SAMOVAR, Télécom SudParis,\\
    Institut Polytechnique de Paris,\\
    91120 Palaiseau, France \\
  \texttt{faiz.taleb@edf.fr} \\
   \And
 Maryline LAURENT  \\
  SAMOVAR, Télécom SudParis,\\
    Institut Polytechnique de Paris,\\
  \texttt{Maryline.Laurent@telecom-sudparis.eu} \\
  \And
 Ivan GAZEAU \\
  EDF\\
    91120 Palaiseau, France \\
  \texttt{ivan.gazeau@edf.fr} \\
}
\begin{document}
%-------------------------------------------------------------------------------

%don't want date printed
\date{}

\title{ Uncovering Memorization in Time‑Series Imputation Models: LBRM Membership Inference and Its Link to Attribute Leakage
}
\maketitle

\begin{abstract}
Deep learning models for time series imputation are now essential in fields such as healthcare, the Internet of Things (IoT), and finance. However, their deployment raises critical privacy concerns. Beyond the well-known issue of unintended memorization, which has been extensively studied in generative models, we demonstrate that time series models are vulnerable to inference attacks in a black-box setting. In this work, we introduce a two-stage attack framework comprising: (1) a novel membership inference attack based on a reference model that improves detection accuracy, even for models robust to overfitting-based attacks, and (2) the first attribute inference attack that predicts sensitive characteristics of the training data for timeseries imputation model. We evaluate these attacks on attention-based and autoencoder architectures in two scenarios: models that are trained from scratch, and fine-tuned models where the adversary has access to the initial weights. Our experimental results demonstrate that the proposed membership attack retrieves a significant portion of the training data with a tpr@top25\% score significantly higher than a naive attack baseline.
We show that our membership attack also provides a good insight of whether attribute inference will work (with a precision of 90\% instead of 78\% in the genral case).
\end{abstract}


%-------------------------------------------------------------------------------

\section{Introduction}

The memorization problem in generative AI has been widely studied in recent years. Carlini et al.~\cite{carlini_extracting_2021} first introduced the concept in the context of large language models (LLMs), showing how GPT-2 unintentionally memorizes small portions of its training data. This phenomenon, later termed \textit{Unintended Memorization}~\cite{carlini_secret_2019}, raises significant privacy and confidentiality concerns, as generative models may memorize sensitive data which could be exploited by malicious actorsto extract private information.

Privacy attacks aim to infer or extract sensitive information from trained models without direct access to the original data. Notable examples of these attacks include, inference attacks such as \textit{membership inference attack (MIA)} and \textit{attribute inference attack (AIA)} are particularly notable. While MIA determines whether a specific record was part of the training set, AIA goes further by attempting to recover internal characteristics of the data, such as specific features or attributes. In time series, this may include detecting local peaks, lows, or structural patterns within a masked window. This demonstrates that privacy leakage can extend beyond membership, revealing deeper internal structure of the underlying signals. Although MIA has been extensively studied~\cite{liu_encodermi_2021,fu_practical_2023,wu_adapting_2021,he_node-level_2021,carlini_membership_2022}, AIA remains underexplored in the context of time series models.

Time series imputation and forecasting are critical tasks in domains such as IoT systems, healthcare, finance, and transportation. Missing values arise due to sensor failures, irregular sampling, privacy constraints, or communication issues, and their accurate reconstruction is essential for downstream analysis. Deep learning-based imputation methods—such as RNNs, attention-based models, Autoencoders (AEs), and Generative Adversarial Networks (GANs)—have shown strong performance~\cite{wang_deep_2025}. However, these models may inadvertently memorize temporal patterns from training data, creating privacy risks during deployment.

In this work, we introduce a new membership inference attack for time-series imputation models based on a Loss-Based Reference Model (LBRM), designed to exploid unintended memorization in a black-box setting. The core contribution of our study lies in this novel MIA mechanism: by comparing the reconstruction behavior of a target model with that of a carefully matched reference model, our attack isolates memorization effects that are invisible to classical loss-based or overfitting-driven strategies. This enables the detection of training samples with significantly higher accuracy across diverse imputation architectures.

Building on the memorization patterns exposed by LBRM, we also tested an  attribute inference attack (AIA) that focuses on determining whether a peak occurs within a masked segment of the time series.

A key contribution of our study is to demonstrate that MIA and AIA are not independent attack vectors but rather correlated manifestations of the same underlying memorization phenomenon. Specifically, our MIA signal aligns with the effectiveness of the corresponding AIA, meaning that an adversary who recovers a strong MIA curve using our attack would gain higher confidence in launching more sophisticated AIA-based attacks than they would from a “perfect” MIA that merely returns membership labels without reflecting true memorization behavior.
Through extensive experiments, we show a strong empirical relationship between a model’s tendency to memorize particular temporal segments (as revealed by MIA) and its tendency to leak structural attributes of those same segments (as exploited by AIA). In practice, samples that produce high MIA confidence scores are precisely those for which attribute inference becomes substantially more accurate. This correlation suggests that attribute leakage often emerges when the model overfits local temporal patterns. Consequently, our MIA can serve as a predictor or early warning signal for potential AIA vulnerability, providing a principled way to assess when attribute-level privacy risks are likely to arise.

This paper contributes two advances that fill critical gaps in the literature:

\begin{enumerate}
    \item \textbf{A novel reference-based membership inference attack (LBRM).}  
    Unlike self-calibrated MIA methods such as SPV-MIA\cite{fu_practical_2023}, our approach introduces an external reference model with matched reconstruction ability. Membership is inferred by comparing reconstruction losses (DTW) of the target and reference models, providing a reliable indication of memorization in time-series reconstruction tasks.

    \item \textbf{The first attribute inference attacks for time-series imputation models.}  
    We introduce a structural AIA framework capable of inferring peaks features within masked temporal windows. 
    
    \item \textbf{An empirical demonstration of the correlation between MIA and AIA.}  
    We show that LBRM scores strongly correlate with AIA performance across multiple architectures, indicating that our MIA does not only reveal membership leakage but can also serve as an early-warning metric for detecting samples at risk of more severe attacks such as attribute inference. 
\end{enumerate}

Together, these contributions establish a new line of inquiry for privacy analysis in time-series modeling. Our findings reveal that unintended memorization not only leads to membership leakage but also amplifies structural attribute leakage, highlighting the need for stronger privacy auditing tools for time-series imputation models.

\paragraph{Paper Organization.}
Section~\ref{sec:related_work} reviews related work on memorization and inference attacks.  
Section~\ref{sec:mia_method} presents our LBRM membership inference attack and reports our experimental evaluation of MIA.
Section~\ref{sec:aia_method} describes the attribute inference methodology, evaluates AIA performance and analyzes its correlation with MIA.  
Finally, Section~\ref{sec:conclusion} concludes the paper..

\section{Related Work}\label{sec:related_work}

The privacy risks associated with machine learning, particularly in the context of generative and large language models (LLMs), have been extensively studied. Early work by Carlini et al.~\cite{carlini_secret_2019,carlini_extracting_2021} introduced the notion of \emph{unintended memorization}, showing that models trained on sensitive data may reveal private information during the generation process. More recent studies have shown that large-scale extraction attacks on LLMs are possible even under restricted black-box access~\cite{nasr_scalable_2023}, highlighting the persistence of memorization in modern architectures.

Membership inference attacks (MIAs) aim to determine whether a given sample was part of a model’s training dataset. The seminal work of Shokri et al.\ introduced shadow-model attacks for classifiers, which were later extended to generative settings~\cite{hayes_logan_2018,hilprecht_monte_2019}. Carlini et al.~\cite{carlini_membership_2022} proposed LiRA, emphasizing evaluation at low false-positive rates as a realistic operating regime. More recent MIA techniques, such as EncoderMI~\cite{liu_encodermi_2021} and SPV-MIA~\cite{fu_practical_2023}, leverage loss trajectories or internal probabilistic variations to refine inference accuracy.

%Ce paragraphe  devrait être fusionné avec le précédent : tu expliques les travaux effectués puis tu expliques pourquoi ils ne répondent pas à ta problématique (à savoir c'est pas des times séries) et du coup remplacer "existing" par "these"
However, {\bf existing} approaches primarily target images, text, or language models, in which memorization often manifests as overconfident logits or near-verbatim reproduction. Time-series imputation models operate differently: reconstruction is driven by temporal structure and contextual interpolation rather than token-level likelihoods. This renders traditional MIA signals weaker or unreliable. 

% Tu mets la charrue avant les boeufs ! 
%There are other attacks that use reference models for MIA ... Fu and al ... compared to them we use external ... 
{\bf Our work overcomes this limitation by introducing a \emph{Loss-Based Reference Model (LBRM)}.} Unlike the self-calibration logic employed by Fu et al.~\cite{fu_practical_2023}, in which the model generates prompts internally to estimate prediction stability, our approach relies on an \emph{external} reference model serving as a non-memorizing baseline. Membership is inferred by comparing reconstruction discrepancies between the target model and this reference, which proves to be significantly more effective for time-series imputation models, where memorization appears in subtle reconstruction differences rather than explicit confidence scores.
% Et là il faudrait que tu argumentes avec les résultats que tu obtiens ou par un autre papier 
%Tu pourrais aussi évoquer la facilité à mettre en place l'attaque par rapport au papier où il faut plusieurs apprentissages.
%Et surtout rappeler qu'aucun de ces papiers ne cherchent à savoir s'il peuvent aller plus loin que la MIA avec leur attack (et si tu as des arguments pour dire que ça serait difficile pour eux de le faire c'est encore mieux).

Attribute inference attacks (AIAs) aim to recover sensitive information about training samples that goes beyond mere membership. Previous studies have examined AIA in classifiers and graph neural networks (GNNs), demonstrating that structural properties can be leveraged to infer hidden node attributes~\cite{he_node-level_2021,wu_adapting_2021}. Salem et al.~\cite{salem_ml_privacy_2019} explored AIA in healthcare and demographic datasets, while Jia and Gong~\cite{jia_attriguard_2018} proposed AttriGuard as a defense mechanism involving adversarial perturbations. Jayaraman and Evans~\cite{jayaraman_aia_imputation_2022} investigated conceptual links between missing-data imputation and attribute inference. More recent works such as Zhao et al.~\cite{zhao_attribute_inference_2021} and Kabir et al.~\cite{kabir_disparate_privacy_2025} further examined the feasibility of these attacks and the disparate vulnerabilities observed across populations and model families.

Despite this progress, no previous studies have examined AIAs in the context of \emph{deep time-series imputation models}. Furthermore, existing AIA techniques typically focus on  categorical or demographic attributes, whereas time-series settings involve inherently \emph{structural} attributes, such as temporal peaks or other event-like features.
% C'est un beau claim, dommage que tu n'en parles pas aussi bien quand tu décris ta deuxième contribution (le fait que la nature de l'attribut inféré est nouvelle)
Our work is the first to demonstrate that such structural patterns can be inferred in a fully black-box setting by exploiting the reconstruction behavior of modern imputation architectures.

\section{A new Membership Inference Attack based on memorization} \label{sec:mia_method}
% Dire que nous essayons d'exploiter la faille de mémorization pour faire une MIA
% L'idée est de d'essayer de faire une MIA sur des données qui ont été mémorisées par le modèle préalablement
This section \ref{sec:mia_method} details our Loss-Based with Reference Model (LBRM) algorithm, designed to perform membership inference attacks on generative and predictive models. By comparing the performance of a target model with a reference model, our approach identifies memorized data with high accuracy. We describe the algorithm's steps, the construction of the reference model, and the evaluation process to determine membership status.

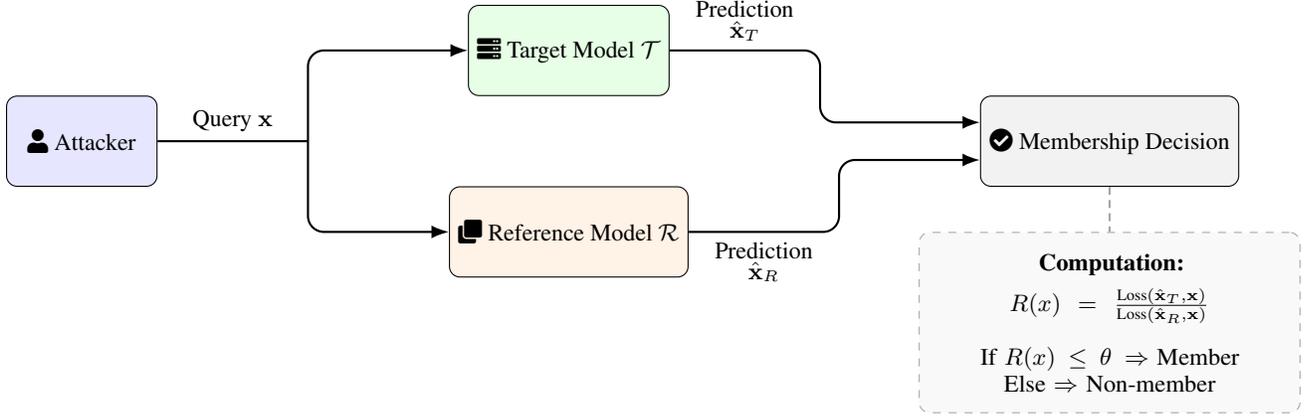
\begin{figure*}[t]
    \centering
    \begin{tikzpicture}[
        font=\small,
        >=Latex, % Sleeker arrowheads
        box/.style={draw, rounded corners, minimum width=2cm, minimum height=1.2cm, align=center},
        lbl/.style={font=\footnotesize, align=center},
        attacker/.style={box, fill=blue!10},
        target/.style={box, fill=green!10},
        reference/.style={box, fill=orange!10},
        decision/.style={box, fill=gray!10},
        arr/.style={->, thick, rounded corners=6pt}
    ]

    % --- 1. Central Models ---
    \node[target] (target) {\faServer \ Target Model $\mathcal{T}$};
    \node[reference, below=1.2cm of target] (reference) {\faClone \ Reference Model $\mathcal{R}$};

    % --- 2. Attacker (Left) & Decision (Right) ---
    % Calculate midpoints for perfect vertical alignment
    \coordinate (mid_left) at ($(target.west)!0.5!(reference.west)$);
    \coordinate (mid_right) at ($(target.east)!0.5!(reference.east)$);

    \node[attacker, left=4cm of mid_left] (attacker) {\faUser \ Attacker};
    \node[decision, right=4cm of mid_right] (decision) {\faCheckCircle \ Membership Decision};

    % --- 3. Math / Computation Block ---
    \node[below=0.6cm of decision, text width=4.5cm, align=center, draw=gray!70, dashed, rounded corners, inner sep=8pt, fill=gray!5] (math) {
        \textbf{Computation:}\\[4pt]
        $R(x) = \frac{\text{Loss}(\hat{\mathbf{x}}_T, \mathbf{x})}{\text{Loss}(\hat{\mathbf{x}}_R, \mathbf{x})}$\\[8pt]
        If $R(x) \leq \theta \Rightarrow$ Member\\
        Else $\Rightarrow$ Non-member
    };
    % Visual link between decision and the math rule
    \draw[densely dashed, gray!70, thick] (decision.south) -- (math.north);

    % --- 4. Routing Arrows ---
    
    % Left Side: Single query forks into two
    \draw[thick] (attacker.east) -- node[above, lbl] {Query $\mathbf{x}$} ++(2,0) coordinate (split);
    \draw[arr] (split) |- (target.west);
    \draw[arr] (split) |- (reference.west);

    % Right Side: Two distinct predictions enter the decision block
    \coordinate (inT) at ([yshift=0.25cm]decision.west);
    \coordinate (inR) at ([yshift=-0.25cm]decision.west);

    \draw[arr] (target.east) -- node[above, lbl] {Prediction\\[-2pt] $\hat{\mathbf{x}}_T$} ++(2,0) |- (inT);
    \draw[arr] (reference.east) -- node[below, lbl] {Prediction\\[-2pt] $\hat{\mathbf{x}}_R$} ++(2,0) |- (inR);

    \end{tikzpicture}
    \caption{Illustrative overview of the LBRM attack pipeline: the attacker compares predictions from the target and reference models to infer membership.}
    \label{fig:lbrm_overview}
\end{figure*}

Our designed \textit{Loss-Based with Reference Model algorithm (LBRM)}  aims to exploit the phenomenon of memorization by the model during the training process to perform a membership inference attack. This attack is performed by comparing the performance of a target  Model (Model \targetmodel) with that of a reference model (Model \referencemodel). The algorithm takes as input the Model \targetmodel  to be attacked, the Model \referencemodel, the suspicious data $x$ and a threshold \threshold. As output, it returns the classification of the data $x$ as member or not.
For more detailed information, see Algorithm \ref{Algo1} and Figure \ref{fig:lbrm_overview}, which describes the algorithm.

\begin{algorithm}[H]
\caption{Loss-Based with Reference Model Algorithms}
\label{Algo1}
\begin{algorithmic}[1]
\STATE \textbf{Input:}   \( \text{x} \), Model \targetmodel, Model \referencemodel, threshold \threshold, masked unit width $U_{width}$.
\STATE \textbf{Output:} Class of x

\STATE \textbf{Steps:}
\STATE \(x_{masked}\)=\( x \) with $U_{width}$ information masked.
\STATE Use Model \targetmodel and Model \referencemodel to predict/generate data from  \(x_{masked}\).  $\hat{y}_t$ = \targetmodel(\(x_{masked}\)) and $\hat{y}_r$ = \referencemodel(\(x_{masked}\))
\STATE Calculate the loss \( L_T \) between the predicted/generated data by Model \targetmodel and the original data.  \( L_T(x) \)=${Loss}_{\text{DTW}}(\hat{y}_t, x) $ 
\STATE Calculate the loss \( L_R(x) \) between the predicted/generated data by Model \referencemodel and the original data. \( L_R (x)\)=${Loss}_{\text{DTW}}(\hat{y}_r, x) $
\STATE Compute the ratio \( R(x) = \frac{L_T(x)}{L_R(x)} \).
\STATE Return 1 if  \( R \leq \theta \) else return 0.
\end{algorithmic}
\end{algorithm}

%A quel endroit tu spécifies la largeur de l'unité masqué et où elle se situe dans ta CdC quand tu expliques ton protocole expérimental ?

\subsection{Constructing the Reference Model for Benchmarking}
One of the key innovations introduced is the concept of a reference model, named as Model \referencemodel. The primary function of Model \referencemodel is to serve as a benchmark for comparing the outputs of Model \targetmodel with those of an unbiased model. This comparison helps to determine whether Model \targetmodel is making accurate predictions or is merely memorizing the training data. In particular, if both models have similar outputs, it indicates that the prediction is likely genuine. Conversely, a significant discrepancy between the predictions suggests a high probability that the data point $x$ is memorized and comes from the training set of the Model \targetmodel.

To ensure the effectiveness of the attack, it is essential that Model \targetmodel and Model \referencemodel have equivalent performance. One method to validate this equivalence is to evaluate the performance of the models on publicly available datasets.

The construction of Model \referencemodel depends on the specific attack scenario. The closer Model \referencemodel is to Model \targetmodel, the greater the likelihood of a successful attack.

The attacker can use a publicly available time series imputation model for Model \referencemodel. This method uses existing, well-validated models as a reference, ensuring that Model \referencemodel provides an unbiased benchmark for comparison.

Alternatively, if the attacker has some meta-information about the target model, they can attempt to train a similar model themselves using a public dataset. It is crucial that the attacker ensures the trained model achieves equivalent performance to Model \targetmodel for the attack to succeed.

\subsection{Evaluating Predictions and Determining Membership}\label{R-score}
After defining Model \referencemodel, we proceed to the prediction and evaluation phase to compute the ratio $R$. This phase includes the processing of the predictions from both Model \targetmodel and Model \referencemodel and compares their outputs to determine the reliability of the predictions.

For  data $x$, the predicted outputs by Model \targetmodel and Model \referencemodel are denoted as $\hat{y}_t$ and $\hat{y}_r$, respectively. These predictions are then used to calculate the loss functions for both models.

 The loss function for the attack is different then the one used in the training of the models. In our case, the loss function to compute $L_T(x)$ and $L_R(x)$ is Dynamic Time Warping (DTW) similarity. The DTW similarity  function measures the similarity between two time series by aligning them in such a way that the distance between the corresponding points is minimized. The formula for the DTW loss is given by the following equation:
\[ \text{Loss}_{\text{DTW}}(y_1, y_2) = \min \left( \sum_{i=1}^{N} \sum_{j=1}^{M} d(y_1(i), y_2(j)) \right) \]
where $d(y_1(i), y_2(j))$ is the distance between the $i$-th point of $y_1$ and the $j$-th point of $y_2$, and the minimization is taken over all possible alignments of the two time series.\\ 

The ratio $R$ of the losses is computed as follows:
\[ R (x)= \frac{L_T(x)}{L_R(x)} \]

A threshold value $\theta$ is defined to classify the data:
\[ \text{Classify } x \text{ as a member of the training data if } R(x) \leq \theta \]

\subsection{Definition of Threshold \threshold }
This threshold $\theta$ is a critical parameter that helps distinguish between genuine predictions and memorized data. If the ratio $R(x)$ exceeds the threshold, it indicates that the data point $x$ is likely to be part of the training set, suggesting that Model \targetmodel has memorized this data point rather than making a genuine prediction.

There are several ways to define the threshold (\threshold). One approach is to take a set of data that the malicious user knows is not used by the target model (\targetmodel) and compute the mean of the score $R$ of the test set. Then, define \threshold as the mean of the test data plus n times the standard deviation:
\[\theta = \text{mean}(R(x_{test})) + n \times \text{std}(R(x_{test})) \]

Another approach is to take the top $n\%$ of the \( R(x) \) score. This method can be nuanced by the fact that the attacker must have an approximate knowledge of the number of data points belonging to the dataset.

\subsection{Experimentation Methodology}
To evaluate our attack algorithm, we use two time series imputation architectures: SAITS \cite{du_saits_2023}, Transformer \cite{vaswani2017attention} and iTransformer \cite{liu2023itransformer}  which are predictive, and an autoencoder \cite{fu_filling_2024}, which is generative. We chose these architectures to leverage their distinct approaches to handling missing values.\\

\subsubsection{Metrics.} We use the same metrics as in \cite{carlini_membership_2022,liu_membership_2022} to summarize the risk and success probability of the attack. The evaluation metrics are:

\begin{itemize}
\item \textbf{TPR at Low FPR.} Specifies the true positive rate when the false positive rate is fixed at 0.1, highlighting the attack's precision under critical conditions.
 \item \textbf{ROC Curve.} Compares the ratio of true positives to false positives across different thresholds, providing a visual representation of the model's performance.
\item \textbf{AUROC Score.} AUROC is the area under the ROC curves. Measures the model's ability to distinguish between members and non-members of the training set across all thresholds. 

\end{itemize}

\subsubsection{Datasets.} We utilize two datasets of electric consumption patterns: the London SmartMeter Energy Consumption Data \cite{noauthor_smartmeter_nodate} to evaluate SAITS \cite{du_saits_2023} and the ASHRAE dataset \cite{noauthor_ashrae_nodate} to evaluate the autoencoder \cite{fu_filling_2024}. 

\begin{itemize}
    \item London smart SmartMeter Energy Consumption Data  \\   (LSMEC) \cite{noauthor_smartmeter_nodate} is a data set that provide timeseries of energy consumption. Energy consumption readings were collected from 5,567 London households as part of the Low Carbon London project (Nov 2011 - Feb 2014). Readings were taken every half hour. The dataset includes energy consumption (kWh per half hour), household ID, date, and time, for a total of approximately 167 million rows.
    \item   ASHRAE dataset \cite{noauthor_ashrae_nodate} contains one year of hourly point which is about 8784 point meter readings from 1479 buildings across various sites worldwide. It contains energy consumption readings for four energy types: electricity, chilled water, steam, and hot water. The dataset also includes building metadata and weather data to support the development of counterfactual models for assessing energy efficiency improvements.
\end{itemize}

Our evaluation involves comparing the results obtained using our algorithm (LBRM) with those from a naive approach based solely on the loss. The main difference between our approach (LBRM) and the  naive
%C'est pas terrible pour le lecteur d'avoir la présentation de naive Loss après que tu en parles, et je me demande si tu ne peux pas juste la décrire en une formule en disant que ça consiste à utiliser Rnaive(x) = Loss(xT,x)
loss approach lies in how the data is separated. As explained in Section \ref{R-score}, our main contribution lies in our use of the R(x) score to determine membership. In contrast, the naive loss approach only uses L(x) to discriminate between training and test data.

%IVAN : Mieux vendre l'intérêt de la comparaison : il s'agit de montrer que c'est bien le modèle témoin qui permet la réussite de l'attaque ok.
%Rajouter le descriptif de l'attaque : on prend Lp(x) au lieu de R(x).
\subsubsection{Naive loss attack.} For benchmarking purposes, we conducted what we refer to as a Naive Loss Attack. This attack aims to separate training data from test data based solely on the loss, similar to the work presented in \cite{liu_membership_2022}. The underlying principle of this attack is that models tend to have lower loss on training data compared to test data, as they have been optimized specifically on the training set. By leveraging this difference in loss, we can infer whether a particular data point was part of the training set or not.

%IVAN Je ne suis pas sûr que threat model soit le plus adapté, on pourrait juste dire training scenario ? ok
\subsubsection{Experimentation Scenarios.} In our threat models, we envisioned two scenarios: one without fine-tuning and one with fine-tuning. Since our approach requires a reference model (\referencemodel) as input, we considered two realistic scenarios. In the first scenario, the target model is not fine-tuned and is exclusively trained on private data. In the second scenario, the target model is initially trained on public data and then fine-tuned on private data. These two scenarios capture the different contexts in which the target model might operate.\\

We consider a target model \targetmodel and two datasets: a public dataset \publicdataset  and a private dataset \privatedataset. We analyze two scenarios:

\paragraph{Scenario 1:} 
 The target model \targetmodel is trained solely on the private dataset (\privatedataset). We ensure that the \privatedataset and \publicdataset are from different distributions. This distinction is crucial to demonstrate that the attack does not require access to data similar to that of the model \targetmodel .

\paragraph{Scenario 2 (fine-tuning):} 
 We assume a target model \targetmodel initially trained with the public dataset (\publicdataset) and subsequently fine-tuned with the private dataset (\privatedataset). 
\\

In both scenarios the model is accessible via an API for imputing missing values. A malicious user aims to extract private data using our LBRM algorithm, having access to the public dataset and some knowledge of the model's architecture.

\subsubsection{Training models \targetmodel and \referencemodel.}
For experimental purposes, we assume that the attacker has access to the meta-parameters and trains their model using the same methodologies. However, it is important to note that the attack can still be effective even if the models are based on different architectures. The only requirement is that the reference model and the target model have comparable performance levels. Both the target model (\targetmodel) and the reference model (\referencemodel) are trained.

\paragraph{Attention based  Architecture (SAITS, Transformers and iTransformer):}\label{par:model_arch}
The  architecture is configured with a single feature (electricity consumption). It comprises two layers, each with a model dimension of 512. The feed-forward layers have a dimension of 128, and the model employs four attention heads, each with key and value dimensions of 128. No dropout is applied. The model is trained for 100 epochs on a GPU, with a batch size of 64.

\paragraph{Autoencoder Architecture:}
The Autoencoder architecture is designed to handle temporal data. The model comprises an encoder with two layers and a decoder with two layers, both connected through a fully-connected layer at the bottleneck. Similar to the SAITS architecture, it is trained for 100 epochs on a GPU without dropout.\\

\paragraph{Data preparation:} each dataset is divided randomly into public, private, and test sets. The models \targetmodel are trained on the public dataset \publicdataset combined with the private dataset \privatedataset. The models \referencemodel are trained solely on the public dataset \publicdataset. In Table \ref{tab:data_preparation}, we summarize the division of each dataset. The division ratios are as follows:
\begin{itemize}
\item \textit{Scenario 1}: In this scenario, we aim to train the \referencemodel on a different data distribution than the \targetmodel. For this purpose, we randomly divide the data into 2/5 for the \publicdataset and 2/5 for the \privatedataset, and reserve 1/5 for testing. The division for each dataset is as follows: 
\begin{itemize} 
    \item \textbf{LSMEC}: Both the \publicdataset and \privatedataset contain 2226 time series each, while the test dataset has 1113 time series. 
    \item \textbf{ASHRAE}: Both the \publicdataset and \privatedataset contain 590 time series each, while the test dataset has 295 time series. 
\end{itemize}

\item \textit{Scenario 2}: In this scenario, we consider the case of fine-tuning. For this purpose, we divide the data such that 3/5 is used for the \publicdataset, 1/5 for the \privatedataset, and 1/5 for the test dataset. The division for each dataset is as follows:
\begin{itemize}
    \item \textbf{LSMEC}: The \publicdataset contains 3340 time series, while the \privatedataset and test datasets each contain 1113 time series.
    \item \textbf{ASHRAE}: The \publicdataset contains 887 time series, while the \privatedataset and test datasets each contain 295 time series.
\end{itemize}
\end{itemize}

\paragraph{Experimentation setup}\label{experimentation_setup}
We train all models on one month of time-series data, with measurements recorded every 30 minutes, resulting in sequences of length $T = 1440$ (i.e., $30$ days $\times$ $48$ time steps per day). For each query, we remove one unit of information at a random temporal position in the sequence, ensuring that the masked location varies across samples. Because a full day corresponds to 48 half-hour intervals, we set the masking window to $U_{\text{width}} = 48$, which allows us to evaluate the model’s ability to reconstruct a complete daily pattern around the missing segment.

\begin{table}[H]
\centering
\resizebox{0.7\columnwidth}{!}{%
\begin{tabular}{@{}ccccc@{}}
\toprule
 & \multicolumn{2}{c}{Scenario 1} & \multicolumn{2}{c}{Scenario 2} \\ \midrule
\multicolumn{1}{c|}{} & \multicolumn{1}{c|}{LSMEC} & \multicolumn{1}{c|}{ASHRAE} & \multicolumn{1}{c|}{LSMEC} & ASHRAE \\ \midrule
\multicolumn{1}{c|}{\textit{\textbf{Public} \publicdataset}} & 2226 & \multicolumn{1}{c|}{590} & 3340 & 887 \\ \midrule
\multicolumn{1}{c|}{\textit{\textbf{Private} \privatedataset}} & 2226 & \multicolumn{1}{c|}{590} & 1113 & 295 \\ \midrule
\multicolumn{1}{c|}{\textit{\textbf{Test}}} & 1113 & \multicolumn{1}{c|}{295} & 1113 & 295 \\ \bottomrule
\end{tabular}%
}
\caption{Division of datasets in different scenarios}
\label{tab:data_preparation}
\end{table}

\subsection{Experimentation Results}

%IVAN trouver un autre mot que result : properties, performance ? 
\subsubsection{Training models performance.}
One input of the LBRM algorithm is to provide a reference model, model \referencemodel. To maximize the success of the attack, we need to ensure that the model \referencemodel has similar performance to the model \targetmodel. This similarity in performance is crucial because the effectiveness of the attack is highly sensitive to the performance parity between the reference and target models. If there is a significant discrepancy in their performance, the attack's success rate may be adversely affected.
Additionally, to minimize bias in our experimentation, we must ensure that the \targetmodel does not have significant issues with overfitting. This precaution helps to maintain the integrity of our experimental results, ensuring that the observed outcomes are due to the attack methodology rather than artefacts of model training.
To evaluate the performance of each model, we randomly remove 20\% of the data and test the resulting model on that data by computing the Mean Absolute Error (MAE). The results are reported in Table~\ref{tab:training_models}.

Table~\ref{tab:training_models} presents the performance evaluation of SAITS, AE, Transformer, and iTransformer models based on MAE across two scenarios. 
Across both scenarios, all four models demonstrate good performance with training and test MAE values that are close to each other, indicating minimal overfitting. Furthermore, the differences between the target and reference models for each architecture are small, confirming that the performance of \referencemodel and \targetmodel remains similar. This ensures that the experimental setup satisfies the conditions required for the LBRM attack, where performance parity between the two models is essential.
\begin{table*}[h]
\centering
\resizebox{\textwidth}{!}{%
\begin{tabular}{c|cc|cc|cc|cc|cc|cc|cc|cc}
\hline
 & \multicolumn{8}{c|}{Scenario 1} & \multicolumn{8}{c}{Scenario 2} \\ \hline
\textbf{MAE} & \multicolumn{2}{c|}{SAITS} & \multicolumn{2}{c|}{Transformer} & \multicolumn{2}{c|}{iTransformer} & \multicolumn{2}{c|}{AE} & \multicolumn{2}{c|}{SAITS} & \multicolumn{2}{c|}{Transformer} & \multicolumn{2}{c|}{iTransformer} & \multicolumn{2}{c}{AE} \\ \hline
 Model &  \targetmodel &  \referencemodel &  \targetmodel &  \referencemodel &  \targetmodel &  \referencemodel &  \targetmodel &  \referencemodel &  \targetmodel &  \referencemodel &  \targetmodel &  \referencemodel &  \targetmodel &  \referencemodel &  \targetmodel &  \referencemodel \\ \hline
Data Train & 0.175 & 0.18 & 0.70 & 0.75 & 0.51 & 0.48 & 0.15 & 0.17 & 0.19 & 0.185 & 0.68 & 0.65 & 0.48 & 0.49 & 0.15 & 0.17 \\ \hline
Data Test  & 0.20  & 0.21 & 0.81 & 0.85 & 0.55 & 0.52 & 0.20 & 0.21 & 0.24 & 0.21  & 0.75 & 0.74 & 0.55 & 0.52 & 0.20 & 0.21 \\ \hline
\end{tabular}%
}
\caption{Performance Evaluation of SAITS, AE, iTransformer, and Transformer Models Based on Mean Absolute Error (MAE)}
\label{tab:training_models}
\end{table*}

\subsubsection{Attack result.}

\begin{table*}[h]
\centering

\resizebox{\textwidth}{!}{%
\begin{tabular}{lcccccccc}
\toprule
\textbf{Metric} & \multicolumn{2}{c}{SAITS} & \multicolumn{2}{c}{iTransformer} & \multicolumn{2}{c}{Transformer} & \multicolumn{2}{c}{AE} \\ \midrule
 & LBRM & Naive Loss & LBRM & Naive Loss & LBRM & Naive Loss & LBRM & Naive Loss \\ \midrule
AUROC & \textbf{0.71} & 0.52 & \textbf{0.65} & 0.53 & \textbf{0.68} & 0.50 & \textbf{0.77} & 0.42 \\
TPR@0.1 & \textbf{0.26} & 0.11 & \textbf{0.19} & 0.09 & \textbf{0.25} & 0.10 & \textbf{0.45} & 0.08 \\
TPR@top25\% & \textbf{60\%} & 52\% & \textbf{55\%} & 40\% & \textbf{60\%} & 50\% & \textbf{72\%} & 40\% \\ \bottomrule
\end{tabular}%
}
\caption{Membership Attack Success Metrics for Scenario 1 (AUROC, TPR@0.1, TPR@top25\%)}

\label{tab:LBRM_result_scenario1}
\end{table*}

% Table for Scenario 2
\begin{table*}[h]
\centering

\resizebox{\textwidth}{!}{%
\begin{tabular}{lcccccccc}
\toprule
\textbf{Metric} & \multicolumn{2}{c}{SAITS} & \multicolumn{2}{c}{iTransformer} & \multicolumn{2}{c}{Transformer} & \multicolumn{2}{c}{AE} \\ \midrule
 & LBRM & Naive Loss & LBRM & Naive Loss & LBRM & Naive Loss & LBRM & Naive Loss \\ \midrule
AUROC & \textbf{0.90} & 0.55 & \textbf{0.85} & 0.53 & \textbf{0.90} & 0.50 & \textbf{0.85} & 0.52 \\
TPR@0.1 & \textbf{0.75} & 0.12 & \textbf{0.72} & 0.10 & \textbf{0.73} & 0.10 & \textbf{0.56} & 0.12 \\
TPR@top25\% & \textbf{92\%} & 48\% & \textbf{85\%} & 40\% & \textbf{90\%} & 50\% & \textbf{80\%} & 51\% \\ \bottomrule
\end{tabular}%
}

\caption{Attack Success Metrics for Scenario 2 (AUROC, TPR@0.1, TPR@top25\%)} 

\label{tab:LBRM_result_scenario2}
\end{table*}

Tables \ref{tab:LBRM_result_scenario1} and \ref{tab:LBRM_result_scenario2} present a comparative analysis of the attack success rates using the proposed LBRM algorithm and the Naive Loss baseline across four architectures: SAITS, AE, Transformer, and iTransformer, evaluated under two experimental scenarios.

In Scenario 1, AUROC values under Naive Loss remain close to random (0.50--0.55), whereas LBRM achieves substantial improvements: SAITS rises from 0.52 to 0.71, AE from 0.42 to 0.77, Transformer from 0.50 to 0.68, and iTransformer from 0.53 to 0.65. Similar trends are observed in Scenario 2, where LBRM reaches AUROC scores of 0.90 for SAITS and Transformer, and 0.85 for iTransformer and AE, compared to Naive Loss values around 0.50--0.55. These gains demonstrate that the attack remains effective regardless of the underlying architecture.

The improvements in TPR@0.1 and TPR@top25\% further highlight the practical impact of LBRM. For example, in Scenario 2, SAITS achieves a TPR@0.1 of 0.75 and a TPR@top25\% of 92\%, while Transformer and iTransformer reach 90\% and 85\%, respectively. Even AE, which is generally considered more robust, shows a TPR@top25\% of 80\%. These high detection rates indicate that the attack remains effective even when restricted to the top-ranked samples, posing a significant privacy risk.
Overall, the results demonstrate that LBRM is architecture-agnostic and scales effectively across different model types. The consistent performance gains over Naive Loss confirm that leveraging reconstruction-based risk scores provides a strong advantage for membership inference attacks. Among the evaluated architectures, SAITS and Transformer exhibit the highest vulnerability, as reflected by their superior AUROC and TPR values under LBRM, suggesting that models with strong temporal or attention mechanisms may be particularly susceptible to this type of attack.

\section{Exploiting Memorization for Attribute Inference Attacks}\label{sec:aia_method}
Although simpler in form, the AIA leverages the same reconstruction behaviour that drives the success of our MIA. To ensure that the attack behaves consistently across the full temporal range, we evaluate its performance using a sliding-window protocol, which allows us to verify the robustness of the method rather than to target any particular region of the signal.
The objective of the proposed attack is to infer the \textbf{presence and temporal location} of sensitive events within an unobserved segment of a time series. Specifically, the attack aims to detect the existence of local extrema (hereafter referred to as \emph{peaks}) occurring inside a missing temporal window.  
Given only partial observations of the signal and black-box access to a trained imputation model, the adversary seeks to determine whether a significant peak occurred within the masked interval by exploiting the reconstruction behavior of the model.

\subsection{Privacy Implications}\label{subsec:aia_motivation}

In the energy sector, time-series data from smart meters (e.g., the LSMEC dataset) serves as a proxy for human behavior. Our structural AIA is specifically designed to uncover "Power Load Signatures" that models may inadvertently memorize. 

The recovery of temporal peaks in energy data is particularly sensitive for two reasons:
\begin{enumerate}
    \item \textbf{Device Profiling:} Individual appliances have distinct load signatures. Accurately imputing a sharp peak allows an adversary to perform Load Monitoring to identify household assets.
    \item \textbf{Behavioral Mapping:} The temporal alignment of energy spikes reveals occupancy patterns. By successfully matching peaks within masked segments, our attack demonstrates that imputation models can leak when a resident is active, thereby facilitating physical security risks or unauthorized surveillance.
\end{enumerate}
\newpage
\subsection{Attack Requirement}
We consider a adversary operating under the following assumptions:
\begin{itemize}
    \item \textbf{Knowledge:} The adversary observes a partially available time series $X_{\text{obs}}$, in which a contiguous temporal window $W$ is unobserved. The adversary has query access to an imputation model $\mathcal{M}$.
    
    \item \textbf{Capabilities:} The adversary can submit masked inputs to $\mathcal{M}$ and obtain reconstructed outputs for the missing values.
    
    \item \textbf{Goal:} Without direct access to the ground truth values within $W$, the adversary aims to infer whether a peak exists in the missing window and, if so, estimate its temporal location.
\end{itemize}

\subsection{Attack Methodology}
%mieux définir W
For a given target window $W$, the attack proceeds as follows:

\begin{enumerate}
    \item \textbf{Query Construction:}  
    The adversary constructs an input $X_{\text{query}}$ by retaining all observed values and replacing the entries corresponding to $W$ with missing-value indicators.

    \item \textbf{Imputation Query:}  
    The query is submitted to the imputation model $\mathcal{M}$, producing a fully reconstructed time series:
    \begin{equation}
        y = \mathcal{M}(X_{\text{query}}).
    \end{equation}

    \item \textbf{Peak Inference:}  
    %faire un paralléle avec l'experimentation mieux définir peak inference par les index 
    The adversary extracts the reconstructed segment $y$ and applies a Continuous Wavelet Transform (CWT)-based peak detection algorithm.  
    The detection of a peak within $y$ is interpreted as evidence of a corresponding peak in the original, unobserved data.
\end{enumerate}

\subsection{Experimental Protocol}

To quantify potential information leakage, we adopt a comparative sliding-window evaluation protocol that contrasts the inference performance of a target model against a baseline model that has not been trained on the sensitive dataset.

\subsubsection{Models Under Comparison}

In the attribut inference attack we stick with the model attention based. we experiment the attack on SAITS\cite{du_saits_2023}, Transformer \cite{vaswani2017attention} and iTransformer \cite{liu2023itransformer}.
\begin{itemize}
    \item \textbf{Target Model ($\mathcal{M}_{\text{target}}$):}  
    The imputation model trained on the private dataset. Elevated inference performance may indicate memorization or leakage of sensitive temporal patterns.
    
    \item \textbf{Evaluation Model ($\mathcal{M}_{\text{eval}}$):}  
    A baseline model trained on disjoint data set than the Target model . This model captures reconstruction behavior driven solely by general statistical regularities.
\end{itemize}

Both models use exactly the \textbf{same hyperparameters, training configuration, and architectural settings} to ensure that the only difference between them is \emph{their training data}.  We use the same architecture as in \ref{par:model_arch}.
This setup allows isolating whether the target model exhibits 
memorization-driven leakage: if the target model consistently outperforms the 
evaluation model on the attribute inference task, this suggests that it has 
internalized specific temporal patterns from sensitive data.

\subsubsection{Peak Detection via Continuous Wavelet Transform}

Peak detection within each temporal window is performed using a Continuous Wavelet Transform (CWT)-based approach. Specifically, we employ the peak detection method implemented in the \texttt{scipy.signal} library, which identifies local extrema by analyzing the signal across multiple temporal scales.

Let $x = (x_1, \dots, x_T)$ denote a discrete time series. The Continuous Wavelet Transform of $x$ with respect to a wavelet $\psi$ is defined as:
\[
\mathcal{W}_\psi(x)(a, b) = \sum_{t=1}^{T} x_t \, \psi\!\left(\frac{t - b}{a}\right),
\]
where $a > 0$ denotes the scale parameter and $b$ the translation parameter.

The CWT-based peak detection algorithm operates by computing wavelet coefficients across a predefined set of scales and identifying ridge-like structures corresponding to local maxima that persist across multiple scales. Peaks are defined as time indices that exhibit consistent local prominence across these scales, which provides robustness to noise and minor temporal variations.

Formally, given a signal segment restricted to window $I_t$, the peak detection function is defined as a mapping:
\[
\mathcal{D}_{\text{CWT}} : \mathbb{R}^{|I_t|} \rightarrow 2^{I_t},
\]
where $\mathcal{D}_{\text{CWT}}(x_{I_t})$ returns a set of temporal indices corresponding to detected peaks within the window.

The output of the detector is therefore a finite set:
\[
P(t) = \mathcal{D}_{\text{CWT}}(x_{I_t}) \subset I_t,
\]
where each index $\tau \in P(t)$ corresponds to a time step at which a local extremum is detected.

By leveraging multi-scale analysis, the CWT-based detector is less sensitive to isolated fluctuations and emphasizes structurally significant peaks. This property is particularly well-suited for the attack setting considered in this work, where reconstruction noise and smoothing effects introduced by the imputation model may distort the original signal.

\subsubsection{Sliding-Window Procedure}

To systematically assess the privacy risks associated with imputation, we employ a 
sliding-window protocol that probes every temporal region of the time series. 
The goal is twofold: (i) evaluate how the attack performs across different masked 
intervals, thereby challenging the robustness of the imputation models, and 
(ii) identify whether certain timestamps are more susceptible to memorization 
than others.  
This protocol is applied independently to both the target model and the evaluation 
model, enabling a comparative analysis of potential memorization-driven leakage.

Let $X \in \mathbb{R}^{N \times T}$ denote the complete dataset. We define a window length $W = 24$ and stride $S = 24$. For each window interval $I_t = [t, t+W)$, the following steps are performed:

\begin{enumerate}
    \item \textbf{Masking:}  
    A masked signal $X_{\text{masked}}$ is created by replacing all values within $I_t$ with missing-value indicators, while preserving the surrounding temporal context.

    \item \textbf{Reconstruction:}  
    the target model ${M}_{\text{target}}$ is queried using the masked input:
    \[
        y^{\text{tgt}} = \mathcal{M}_{\text{target}}(X_{\text{masked}}), \qquad
    \]
    \fancyfoot[R]{gt: ground truth;
    tgt: target}
    \item \textbf{Window-Restricted Peak Detection:}  
    For each window $I_t = [t, t+W)$, peak detection is applied independently to the ground truth signal $X$ and to the reconstructed signals.
    Formally:
    \[
        \mathcal{P_{\text{gt}}}(t) \subset I_t, \qquad
        \mathcal{P_{\text{tgt}}}(t) \subset I_t,
    \]
    where each set contains the temporal indices of detected peaks.

    \item \textbf{Attack Success Criterion:}  
   We define the attack as a pointwise classification task: for each timestamp $t$, 
    the adversary predicts whether $t$ corresponds to a peak in the underlying 
    (time‑masked) ground truth signal.

Formally, the predicted class for index $t$ in series $i$ is:
\[
    \hat{c}_i^t =
    \begin{cases}
        1 & \text{if } \exists\, p_{\text{imp}} \in \mathcal{P}_i^{imp} 
        \text{ such that } |p_{\text{imp}} - t| \le \tau, \\
        0 & \text{otherwise}.
    \end{cases}
\]

The model’s prediction is considered \emph{correct} if it matches the ground‑truth 
classification:
\[
    c_i^t =
    \begin{cases}
        1 & \text{if } \exists\, p_{\text{gt}} \in \mathcal{P}_i^{gt} 
        \text{ such that } |p_{\text{gt}} - t| \le \tau, \\
        0 & \text{otherwise},
    \end{cases}
\]
where $\tau$ is the temporal tolerance used to relate peaks to their nearby indices.

A timestamp is therefore correctly classified when $\hat{c}_i^t = c_i^t$. 
From these predictions, we derive the standard confusion-matrix terms as follows:

\begin{align}
TP &= \bigl|\{(i,t) \;\mid\; \hat{c}_i^t = 1 \;\wedge\; c_i^t = 1\}\bigr|, \\[4pt]
FP &= \bigl|\{(i,t) \;\mid\; \hat{c}_i^t = 1 \;\wedge\; c_i^t = 0\}\bigr|, \\[4pt]
TN &= \bigl|\{(i,t) \;\mid\; \hat{c}_i^t = 0 \;\wedge\; c_i^t = 0\}\bigr|, \\[4pt]
FN &= \bigl|\{(i,t) \;\mid\; \hat{c}_i^t = 0 \;\wedge\; c_i^t = 1\}\bigr|.
\end{align}

    \item \textbf{Performance Metrics:}  
    %justifier l'utilisation de ces metrics 
    Detected peaks are matched to ground truth peaks using a temporal tolerance $\delta$.  
    The attack performance is quantified using \textbf{precision} and \textbf{recall}. \textbf{Privacy Implication:} High recall  combined with high precision indicates the model has memorized specific peak patterns, representing a Privacy Risk.

    \begin{itemize}
        \item \textbf{Precision} measures the proportion of inferred peaks that correspond to true peaks, reflecting the reliability of the attack.
       
        \[
        \text{Precision} = \frac{TP}{TP + FP} \\
         \]
         
        \item \textbf{Recall} measures the proportion of ground truth peaks within the masked window that are successfully inferred, reflecting the effectiveness of the attack.
        \[\text{Recall} = \frac{TP}{TP + FN} \\\]

    \end{itemize}
\end{enumerate}

\begin{comment}
A peak in a one-dimensional time series is defined, following the 
\texttt{scipy} library, as a sample index 
$t_0$ for which the Continuous Wavelet Transform (CWT) of the signal 
exhibits a relative maximum across multiple wavelet scales. 
Formally, let $x(t)$ be the signal and let $W_x(a,t)$ denote the CWT 
computed using a chosen wavelet (typically the Ricker wavelet). 
The point $t_0$ is identified as a peak if:

\begin{enumerate}
    \item $W_x(a,t_0)$ is a local maximum in $t$ for several scales $a$,
    \item these local maxima form a connected ridge line across adjacent
          scales, and
    \item the ridge line satisfies minimal length and signal-to-noise 
          ratio (SNR) requirements.
\end{enumerate}

Only ridge lines appearing consistently across scale space and meeting 
the filtering criteria (e.g., minimum ridge length, SNR threshold, 
maximum allowed gaps) are retained as valid peaks.%
\cite{scipy_find_peaks_cwt_doc}
\end{comment}
\newpage
\subsubsection{Parameters}

The table \ref{tab:parameters} outlines the configuration parameters employed for peak detection and windowing:

\begin{table}[h]
\centering
\renewcommand{\arraystretch}{1.2} % Balanced spacing for academic style
\begin{tabularx}{\columnwidth}{@{} l c c X @{}}
\toprule
\textbf{Parameter} & \textbf{Sym.} & \textbf{Val.} & \textbf{Description} \\ 
\midrule
Window size  & $w$      & 24       & Consecutive timestamps to mask \\
Stride       & $s$      & 24       & Step size between windows \\
Tolerance    & $\tau$   & 2        & Max distance for peak matching \\
CWT widths   & $\omega$ & $\{1..4\}$ & Continuous Wavelet Transform scales \\ 
\bottomrule
\end{tabularx}
\caption{Summary of experimental parameters.}
\label{tab:parameters}
\end{table}

\subsubsection{Data Distribution}
Since the attribute inference attack primarily targets attention-based architectures, we employ the same dataset used in the MIA experiments: the London SmartMeter Energy Consumption (LSMEC) dataset \cite{noauthor_smartmeter_nodate}. 

In this scenario, the \textbf{target dataset} is restricted to the subset of time series where the MIA attack was successful, as these sequences represent the highest privacy risk. The \textbf{test set} remains identical to that used in the previous experiments to ensure consistency in evaluation.

Unlike the MIA experiments, where we considered two scenarios (with and without fine-tuning), we do not replicate this distinction for the attribute inference attack. The reason is that attribute inference evaluates privacy leakage at a finer granularity—specifically, the ability of the model to memorize and reveal internal characteristics of the data rather than its mere presence in the training set. Fine-tuning primarily affects global memorization patterns and membership detection, but attribute-level inference depends on localized behaviors within the model's learned representations. Therefore, introducing a fine-tuning scenario would not significantly alter the risk assessment for attribute inference, as the attack focuses on structural memorization rather than global adaptation.

\begin{figure*}[!t]
    \centering
    \begin{tikzpicture}[
        font=\small,
        arr/.style={-{Latex}, thick},
        blk/.style={draw, rounded corners, align=center, inner sep=6pt, fill=blue!5},
        sel/.style={draw, rounded corners, align=center, inner sep=6pt, fill=purple!8},
        ds/.style={draw, rounded corners, align=center, inner sep=5pt, fill=gray!10},
        step/.style={draw, rounded corners, inner sep=6pt, align=left, fill=yellow!7}
    ]

    % --- 1. Titre global ---
    \node[align=center, text width=\textwidth] (title) {\textbf{MIA + AIA at Series Level:} LBRM risk on seen part $\rightarrow$ AIA on unseen regions};

    % --- 2. Colonne du milieu (LBRM Pipeline) ---
    % On place le centre en premier pour ancrer le reste
    \node[blk, text width=0.26\textwidth, below=0.8cm of title] (lbrm) {\textbf{LBRM on seen part}\\[2pt] Global risk score $R(x_{\text{seen}})$ per series};
    \node[blk, text width=0.26\textwidth, below=0.5cm of lbrm] (rank) {\textbf{Rank by $R$}\\[2pt] Select Top-$q\%$ (e.g., 25\%)};
    \node[sel, text width=0.26\textwidth, below=0.5cm of rank] (subset) {High-risk set $\mathcal{S}_{\text{risk}}$};

    % --- 3. Colonne de gauche (Attacker View) ---
    % Alignement parfaitement centré verticalement avec le bloc 'rank'
    \node[ds, minimum width=0.30\textwidth, minimum height=4cm, left=0.06\textwidth of rank] (seriesbox) {};

    % Contenu de la seriesbox (positionné de façon relative au centre de la boîte, zéro problème d'alignement)
    \node[anchor=north, font=\bfseries] (att_title) at ([yshift=-0.15cm]seriesbox.north) {Attacker view (per series)};
    \node[anchor=north, font=\scriptsize] (legend) at ([yshift=-0.05cm]att_title.south) {\textcolor{green!40!black}{green} = seen \quad \textcolor{red!60!black}{red} = unseen};

    % Série j (Milieu)
    \node[anchor=east] (s2_seen) at ([xshift=0.2cm, yshift=-0.4cm]seriesbox.center) [draw=green!40!black, fill=green!25, thick, minimum width=1.5cm, minimum height=0.4cm, inner sep=0] {};
    \node[anchor=west] (s2_un) at (s2_seen.east) [draw=red!60!black, fill=red!12, dashed, thick, minimum width=1.5cm, minimum height=0.4cm, inner sep=0] {};
    \node[anchor=east, font=\footnotesize] (s2_name) at ([xshift=-0.15cm]s2_seen.west) {Series $j$};

    % Série i (Haut)
    \node[above=0.25cm of s2_seen, anchor=south] (s1_seen) [draw=green!40!black, fill=green!25, thick, minimum width=1.5cm, minimum height=0.4cm, inner sep=0] {};
    \node[anchor=west] (s1_un) at (s1_seen.east) [draw=red!60!black, fill=red!12, dashed, thick, minimum width=1.5cm, minimum height=0.4cm, inner sep=0] {};
    \node[anchor=east, font=\footnotesize] (s1_name) at ([xshift=-0.15cm]s1_seen.west) {Series $i$};

    % Série k (Bas)
    \node[below=0.25cm of s2_seen, anchor=north] (s3_seen) [draw=green!40!black, fill=green!25, thick, minimum width=1.5cm, minimum height=0.4cm, inner sep=0] {};
    \node[anchor=west] (s3_un) at (s3_seen.east) [draw=red!60!black, fill=red!12, dashed, thick, minimum width=1.5cm, minimum height=0.4cm, inner sep=0] {};
    \node[anchor=east, font=\footnotesize] (s3_name) at ([xshift=-0.15cm]s3_seen.west) {Series $k$};

    % --- 4. Colonne de droite (AIA Pipeline) ---
    % On aligne le haut de AIA avec le haut de LBRM
    \node[step=1, text width=0.28\textwidth, right=0.06\textwidth of lbrm, anchor=north west] (aia) {
        \textbf{AIA on unseen regions of }$\mathcal{S}_{\text{risk}}$:\\[4pt]
        1. Mask a window $W$ in the unseen part\\
        2. Query model $\Rightarrow$ reconstructed $\hat{x}$\\
        3. Detect peaks in $\hat{x}\!\mid_{W}$ (CWT)\\
        4. Measure precision/recall
    };
    \node[blk, fill=orange!10, text width=0.28\textwidth, below=0.5cm of aia] (out) {\textbf{Outcome:} Higher AIA precision on $\mathcal{S}_{\text{risk}}$ compared to all series};

    % --- 5. Flèches et Connexions ---
    \draw[arr] (seriesbox.east |- lbrm.west) -- (lbrm.west);
    \draw[arr] (lbrm.south) -- (rank.north);
    \draw[arr] (rank.south) -- (subset.north);
    \draw[arr] (subset.east) to[out=0, in=180] (aia.west); % Courbe fluide pour lier la sélection à l'AIA
    \draw[arr] (aia.south) -- (out.north);

    \end{tikzpicture}
    \caption{MIA + AIA pipeline. The attacker computes LBRM on the \emph{seen} portion to obtain a series-level risk score, ranks series, selects the Top-$q\%$ high-risk set, and runs AIA on \emph{unseen} regions of those series.}
    \label{fig:mia-aia-full}
\end{figure*}
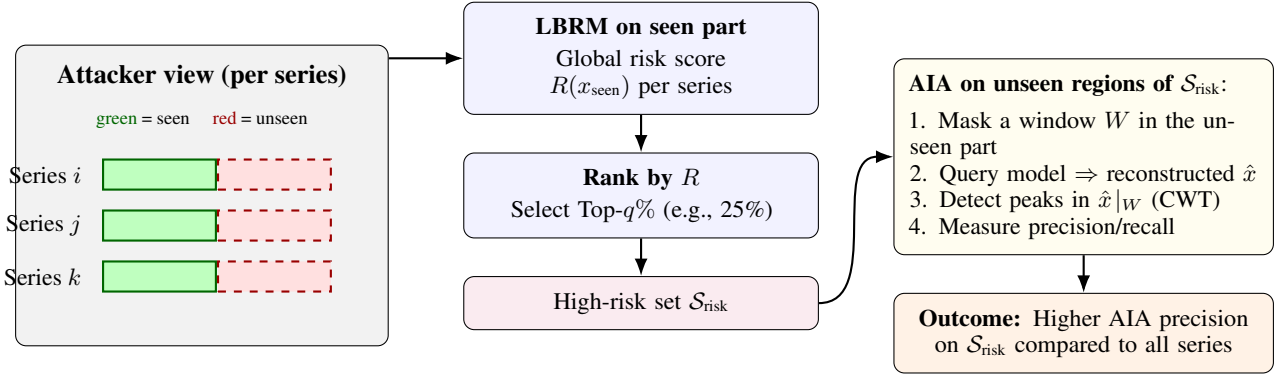

\subsection{Result \& Discussion}
As motivated in Section \ref{subsec:aia_motivation}, the leakage of structural peaks in energy data is not merely a reconstruction error but a direct exposure of household appliance signatures and occupancy patterns.
Across all three architectures, the results consistently reveal a substantial gap between the Target and Evaluation models, indicating that each model memorizes sensitive temporal patterns from the training data to varying degrees. The Transformer exhibits clear leakage with pronounced variability across windows, suggesting that certain temporal regions are particularly vulnerable. SAITS demonstrates more stable and accurate imputation, yet the strong performance gap between its Target and Evaluation versions shows that this robustness is accompanied by significant memorization of training‑specific peak structures. iTransformer achieves the highest reconstruction fidelity overall, and consequently displays the most pronounced leakage, with uniformly high recall and precision gains that suggest deeper internalization of temporal dependencies.
\begin{table}[H]
\centering
\setlength{\tabcolsep}{4pt} % compact column spacing
\renewcommand{\arraystretch}{1.15} % slightly larger row spacing

\begin{tabular}{lcc}
\hline
\textbf{Model} & \textbf{Recall (avg ± std)} & \textbf{Precision (avg ± std)} \\
\hline
\multicolumn{3}{l}{\textbf{Transformer}} \\
\quad Target       & 0.87 ± 0.05 & 0.75 ± 0.06 \\
\quad Evaluation   & 0.48 ± 0.08 & 0.40 ± 0.05 \\
\hline
\multicolumn{3}{l}{\textbf{SAITS}} \\
\quad Target       & 0.90 ± 0.04 & 0.78 ± 0.05 \\
\quad Evaluation   & 0.55 ± 0.07 & 0.43 ± 0.05 \\
\hline
\multicolumn{3}{l}{\textbf{iTransformer}} \\
\quad Target       & 0.92 ± 0.03 & 0.80 ± 0.05 \\
\quad Evaluation   & 0.57 ± 0.06 & 0.45 ± 0.04 \\
\hline
\end{tabular}

\caption{Average and standard deviation of recall and precision across all sliding windows (mask size = 24) for the three attention-based architectures. Target and Evaluation models share identical hyperparameters but are trained on disjoint sample sets.}
\label{tab:slidingwindow-results}
\end{table}

\subsection{Does LBRM--MIA Predict Attribute Leakage in Unseen Regions?}

We now evaluate whether the Loss-Based Reference Model (LBRM) can be used not only to detect membership leakage but also to \emph{predict} which entire time series are globally at risk of attribute inference. Crucially, our goal is not to determine which \emph{windows} are risky, but whether an attacker—who only sees a partial portion of a time series—can use LBRM to assess whether the \emph{unseen regions} of that same time series are more vulnerable to AIA.

In this scenario, the attacker computes the LBRM score on the portion of the sequence they have access to. This score is then treated as a global ``risk indicator’’ for the entire time series. The attacker then performs AIA on regions they do \emph{not} have access to. If LBRM effectively captures memorization, then sequences with higher LBRM scores should exhibit systematically stronger attribute leakage even in the parts that were never observed by the attacker.

To test this hypothesis, we compute the Pearson correlation coefficient~\cite{pearson1895} between the LBRM score (computed only on the part visible to the attacker) and the AIA precision and recall (evaluated on unseen regions of the same sequence). Table~\ref{tab:corr-lbrm-aia} reports the results for SAITS, Transformer, and iTransformer. Across architectures, we observe a strong and statistically significant correlation between LBRM and AIA \emph{precision}. This indicates that time series that appear memorized in their observed portion are precisely those whose unobserved portions leak structural attributes more reliably. Correlations with recall remain weak, suggesting that memorization primarily affects the \emph{confidence} of attribute leakage rather than the frequency with which peaks are detected.

\paragraph{Time-Series-Level Risk Selection (Top‑25\%).}
We further test whether an attacker can use LBRM to \emph{select entire time series} that are globally vulnerable to attribute inference, as shown in Figure~\ref{fig:mia-aia-full}. 
For each sequence, we compute its LBRM score on the observed part by removing a single unit width of 48  at a random position, following the procedure described in Section~\ref{experimentation_setup}. 
The sequences are then ranked according to their LBRM score, and we select the top 25\% most vulnerable ones. 
Attribute inference is subsequently evaluated on an \emph{unseen} region of length 24 time steps, matching the AIA window size. 

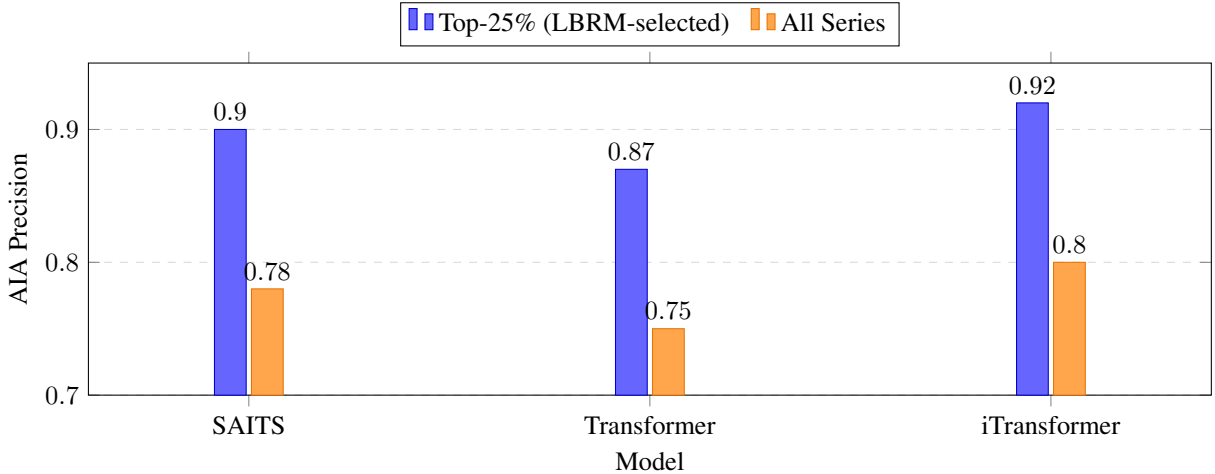
\begin{figure}[t]
\centering
\begin{tikzpicture}
\begin{axis}[
    ybar,
    width=\columnwidth,
    height=6.0cm,
    bar width=12pt,
    ymin=0.70, ymax=0.95,    % tighten to your value range for readability
    enlarge x limits=0.20,
    symbolic x coords={SAITS,Transformer,iTransformer},
    xtick=data,
    ylabel={AIA Precision},
    xlabel={Model},
    legend style={at={(0.5,1.05)},anchor=south,legend columns=-1, /tikz/every even column/.style={column sep=6pt}},
    ymajorgrids=true,
    grid style={dashed,gray!30},
    tick label style={/pgf/number format/fixed},
    nodes near coords,
    nodes near coords align={vertical},
    % Colors (can adjust to match your paper's palette)
    cycle list={{fill=blue!60, draw=blue!80!black}, {fill=orange!70, draw=orange!90!black}}
]
% Top-25% (LBRM-selected)
\addplot coordinates {(SAITS,0.90) (Transformer,0.87) (iTransformer,0.92)};
% All Series
\addplot coordinates {(SAITS,0.78) (Transformer,0.75) (iTransformer,0.80)};

\legend{Top-25\% (LBRM-selected), All Series}
\end{axis}
\end{tikzpicture}
\caption{AIA precision on unseen parts of the time series. ``Top‑25\%'' indicates selecting whole sequences based on their LBRM score computed from the part accessible to the attacker.}
\label{fig:mia-aia-hist}
\end{figure}

Figure~\ref{fig:mia-aia-hist} compares AIA precision between  
(i) all sequences (baseline), and  
(ii) the top--25\% highest-risk sequences identified by LBRM.  

LBRM-based selection consistently boosts AIA precision by 10--15 percentage points across all architectures, showing that LBRM not only exposes memorized regions but also helps identify full time series that are globally susceptible to structural attribute leakage.
\begin{table}[t]
\centering
\small
\setlength{\tabcolsep}{4pt}
\begin{tabular}{lcccc}
\toprule
\textbf{Model} & \textbf{Rec. (r)} & \textbf{p-val} & \textbf{Prec. (r)} & \textbf{p-val} \\
\midrule
Transformer   & -0.10 & 0.73 & 0.57 & 0.03 \\
SAITS         &  0.18 & 0.51 & 0.64 & 0.02 \\
iTransformer  &  0.26 & 0.34 & 0.71 & 0.01 \\
\bottomrule
\end{tabular}
\caption{Correlation between LBRM membership scores (computed on the attacker-visible portion) and AIA performance (evaluated on unseen regions).}
\label{tab:corr-lbrm-aia}
\end{table}

\paragraph{Discussion.}
%These results demonstrate that LBRM provides a reliable \emph{time-series-level} risk {\bf estimator}:
%Je ne sais pas si c'est une bonne idée d'introduire la notion d'estimateur de risque (ça ne fait pas partie de ton narratif, et on se demande pourquoi tu aurais besoin d'une attaque en boite noire pour ça)
%J'aurai plutôt dit un truc du genre : nos résultats confortent que l'attaque de membership est une première étape vers des attaques plus avancée et même que pour notre MIA celle-ci peut aider à cibler et améliorer des attaques d'attributs
% Dernier point : tes résultats laissent entendre que la mémorization sur une zone temporelle est corrélé avec la mémorization de toute la courbe en général, et qu'il n'a pas juste des petits morceaux diffus qui sont mémorisées, c'est important à souligner ici je pense
%sequences that appear memorized in the observed portion leak more structural information in their unobserved portions. This shows that membership-based memorization and structural attribute leakage emerge from the same underlying mechanisms, and that LBRM can act as an early-warning tool for identifying users or sequences at elevated privacy risk.

These results indicate that our membership inference attack is not an isolated mechanism but rather a first step toward more advanced privacy attacks. In particular, samples flagged as members by LBRM tend to be precisely those for which attribute inference becomes significantly more accurate. This shows that the memorization detected on an observed temporal window aligns with broader memorization patterns across the entire sequence, rather than being limited to small isolated fragments. Consequently, LBRM not only reveals whether a sequence has been memorized but also helps target and strengthen subsequent attribute‑inference attacks by identifying the sequences most susceptible to structural leakage.

\section{Conclusion}
\label{sec:conclusion}

This work demonstrates that deep time-series imputation models are vulnerable to substantial privacy leakage, memorizing not only the presence of training samples but also fine grained structural attributes embedded in their temporal patterns. We introduced the \textbf{Loss-Based Reference Model (LBRM)}, a new membership inference attack that lifts detection accuracy from near‑random baselines to AUROC scores as high as 0.90. By comparing a target model with a matched reference model, LBRM exposes subtle reconstruction discrepancies that traditional MIA signals fail to capture.
Beyond membership, we designed an \textbf{Attribute Inference Attack (AIA)} capable of recovering sensitive local structures such as peaks within masked windows. The attack achieves high recall (0.87--0.92) and precision (0.75--0.80) in a black-box setting, confirming that modern imputation architectures internalize detailed temporal patterns from their training data and can inadvertently reveal them.
A key contribution of our study is to establish a direct operational link between these two forms of leakage. While AIA applied to all sequences yields moderate precision, its performance rises sharply—to 0.87--0.92—when restricted to the top 25\% of sequences identified as high-risk by LBRM. This before–after improvement shows that LBRM not only uncovers local memorization but also identifies \textit{which entire time series are globally vulnerable} to attribute inference. The statistically significant correlations observed between LBRM scores and AIA precision further confirm that both leakages stem from the same underlying memorization mechanisms. Importantly, our findings reveal that memorization within a small observed segment is strongly tied to broader memorization across the full sequence, rather than being limited to isolated fragments.
In summary, our results highlight that privacy risks in timeseries imputation models are deeper and more interconnected than previously recognized. LBRM serves as a dual-purpose tool: an effective membership inference method and a powerful mechanism for targeting and amplifying attribute inference attacks by identifying sequences most susceptible to structural leakage. This combined perspective lays the groundwork for developing more robust privacy auditing frameworks and for designing imputation systems that better balance predictive performance with stronger privacy guarantees.

\bibliographystyle{plain}
\bibliography{usenix2024_SOUPS}

%%%%%%%%%%%%%%%%%%%%%%%%%%%%%%%%%%%%%%%%%%%%%%%%%%%%%%%%%%%%%%%%%%%%%%%%%%%%%%%%
\end{document}